\documentclass[11pt,a4paper]{article}
\usepackage[hyperref]{styleAndBib/acl2019}
\usepackage{times}
\usepackage{latexsym}
\usepackage{graphicx}
\graphicspath{ {./} }
\usepackage{url}
\usepackage{xspace}
\usepackage{comment}

\aclfinalcopy % Uncomment this line for the final submission
\newcommand{\sectionRef}[1]{Section \ref{#1}}
\newcommand{\figRef}[1]{Figure  \ref{#1}}

\newcommand{\listItem}[1]{\textit{\textbf{#1}}}

\newcommand{\K}[0]{K\xspace}
\newcommand{\numLogisticRegressionFeatures}[0]{$23$\xspace}
\newcommand{\numClaimMatchingMethodsSelectedForDisplay}[0]{$3$\xspace}

\newcommand{\hmName}[0]{\emph{HM}\xspace}
\newcommand{\lrName}[0]{\emph{LR}\xspace}
\newcommand{\nnName}[0]{\emph{NN}\xspace}

\newcommand{\numMotions}[0]{$200$\xspace}

\newcommand{\numSpeeches}[0]{$400$\xspace}

\newcommand{\fractionOfMotionsInIdebate}[0]{$39\%$\xspace}
\newcommand{\aceTopicCoveragePercentage}[0]{$93.5\%$\xspace}

\newcommand{\claimsPrecision}[0]{$86\%$\xspace}
\newcommand{\averageClaimsPerSpeech}[0]{$12.2$\xspace}

\newcommand{\averageNumberOfSentencePerSpeech}[0]{$29$\xspace}
\newcommand{\averageNumberOfTokensPerSpeech}[0]{$748$\xspace}
\newcommand{\averageSpeechWER}[0]{$7.07\%$\xspace}
\newcommand{\numDebaters}[0]{$9$\xspace}

\newcommand{\numLabeledSpeechLabelPairs}[0]{4,882\xspace}
\newcommand{\idebate}[0]{iDebate\xspace}
\newcommand{\supp}[0]{Appendix\xspace}
\newcommand{\devSetName}[0]{\textit{train}\xspace}
\newcommand{\testSetName}[0]{\textit{test}\xspace}
\newcommand{\splitSizeMotions}[0]{$100$\xspace}
\newcommand{\splitSizeSpeeches}[0]{$200$\xspace}
\newcommand{\devNumSpeechClaimLabels}[0]{2,456\xspace}
\newcommand{\testNumSpeechClaimLabels}[0]{2,426\xspace}

\newcommand{\cohenKappa}[0]{$0.44$\xspace}

\newtheorem{theorem}{Theorem}
\newtheorem{example}[theorem]{Example}

\usepackage{listings}
\title{Towards Effective Rebuttal:\\Listening Comprehension using Corpus-Wide Claim Mining}

\author{
Tamar Lavee\thanks{\ \ These authors equally contributed to this work.} 
\thanks{\ \ The authors list has been updated in this version compared to the one published in the 6th Argument Mining Workshop. When citing this work, please refer to this version.}, 
Matan Orbach$^{*}$, %
Lili Kotlerman, %
Yoav Kantor, 
Shai Gretz, 
Lena Dankin, \\
\textbf{
Shachar Mirkin, %
Michal Jacovi, %
Yonatan Bilu, %
Ranit Aharonov and %
Noam Slonim}  \\
IBM Research
}
\begin{document}
\maketitle

\begin{abstract}
Engaging in a live debate requires, among other things, the ability to effectively rebut arguments claimed by your opponent.
In particular, this requires identifying these arguments.
Here, we suggest doing so by automatically mining claims from a corpus of news articles containing billions of sentences, and searching for them in a given speech. 
This raises the question of whether such claims indeed correspond to those made in spoken speeches.
To this end, we collected a large dataset of $400$ speeches in English discussing $200$ controversial topics, mined claims for each topic, and asked annotators to identify the mined claims mentioned in each speech.
Results show that in the vast majority of speeches debaters indeed make use of such claims.
In addition, we present several baselines for the automatic detection of mined claims in speeches, forming the basis for future work.
All collected data is freely available for research. 

\end{abstract}

\section{Introduction}\label{sec:intro}

Project Debater\footnote{\url{www.research.ibm.com/artificial-intelligence/project-debater}} is a system designed to engage in a full live debate with expert human debaters.
One of the major challenges in such a debate is listening to a several-minute long speech delivered by your opponent, identifying the main arguments, and rebutting them with effective persuasive counter arguments. 
This work focuses on the former, namely, automatically identifying arguments mentioned in opponent speeches.

One of the fundamental capabilities developed in Debater is the automatic mining of claims \citep{Levy14} -- general, concise statements that directly support or contest a given topic -- from a large text corpus.
It allows Debater to present high-quality content supporting its side within its generated speeches.
Our approach utilizes this capability for a different purpose: 
claims mined \emph{from the opposing side} are searched for in a given opponent speech.

The implicit assumption in this approach is that mined claims would be often said by human opponents.
This is far from trivial, since mined content from a large text corpus is not guaranteed to provide enough coverage over arguments made by individual human debaters.
To assess this, we collected a large and varied dataset of recorded speeches discussing controversial topics, along with an annotation specifying which mined claims are mentioned in each speech. 

Annotation results show our approach obtains good coverage, thus making the task of \emph{claim matching} -- automatically identifying given claims in speeches -- interesting in the context of \emph{mined} claims. 
Using the collected data, several claim matching baselines are examined, forming the basis for future work in this direction.

The main contributions of this paper are: 
(i) a recorded dataset of $400$ speeches discussing $200$ controversial topics, along with mined claims for each topic;
(ii) an annotation specifying the claims mentioned in each speech;
(iii) baselines for matching mined claims to speeches.
All collected data is freely available for further research\footnote{\url{https://www.research.ibm.com/haifa/dept/vst/debating_data.shtml}}.

\section{Related Work}

\citet{Mirkin-etal:idebate} recently presented a dataset similar to the one we collected in the context of Machine Listening Comprehension (MLC) over argumentative content.
Instead of using mined claims, they extracted lists of potential arguments from 
\idebate\footnote{\url{https://idebate.org/debatabase}}, a manually curated high-quality database containing arguments for controversial topics. 
A major drawback of such an approach is topic coverage -- any topic not included in the database cannot be handled.
Another limitation is that argument lists from \idebate are short, each typically contains only $3$ or $4$ arguments from each side. 

MLC has been recently gaining attention, and there are several new interesting works and datasets \cite{li2018spokenSquad, lee2018odsqaDataset, unlu2019sqaLectures}. Other tasks are often phrased as a collection of test questions,
which can be multiple choice \cite{Toefl2016, Fang16toeflTreeStructured} or require, for example, identifying an entity mentioned by the speaker 
\cite{Surdeanu06sqa,Comas2010factiodSQA}.

Methods for detecting claims in given texts have been applied to various argumentative domains (e.g. by \citet{MochalesAndMoens11, stab2017parsing, habernal2017argumentationWebDiscourse}).
While such tools may be applied to opponent speeches, a major difference in our setting is that it involves \emph{spoken} rather than \emph{written} language.
Spoken spontaneous speeches often contain disfluencies such as breaks, repetitions, or other irregularities, and therefore claims detected in spoken content are likely to contain them as well.
In addition, since the opponent speech audio is transcribed into text using an Automatic Speech Recognition (ASR) system, its errors propagate to detected claims.
This is a crucial point for Debater -- since a desired rebuttal in live debates typically includes a quote of the argument made by the opponent.
Thus, any single disfluency or ASR error in a detected claim prevents its actual use.

\section{Data}\label{sec:data}

\begin{figure*}[t]
\centering
\includegraphics[trim={0.8cm 6.5cm 1cm 4.6cm},clip,width=\linewidth]{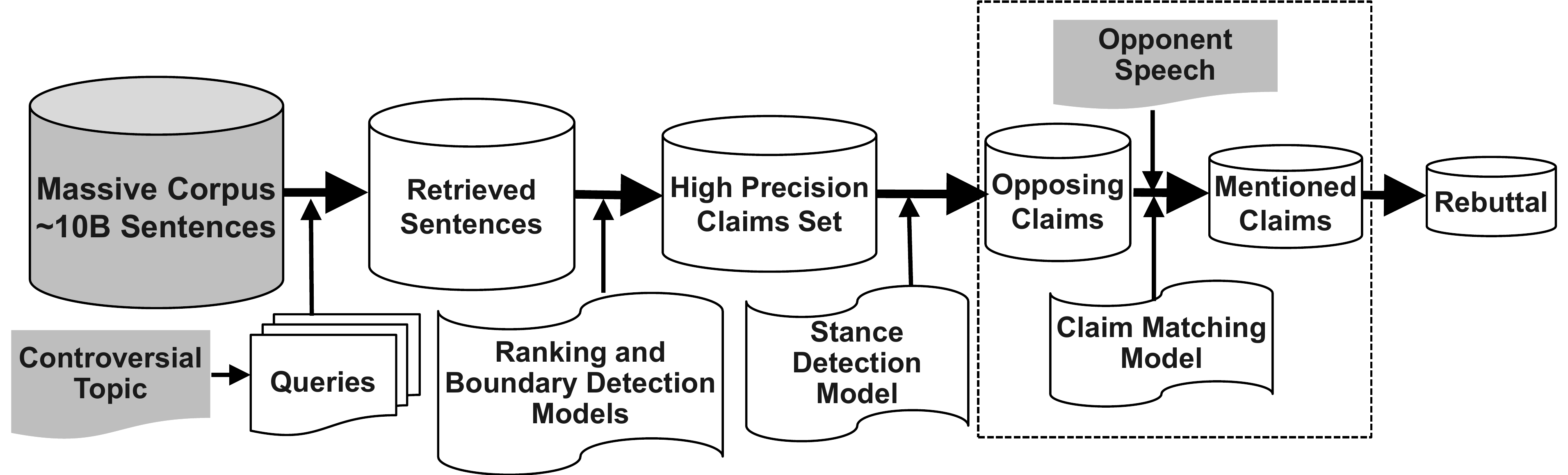}
\caption{The suggested architecture for mined--claims based rebuttal generation.
System inputs are depicted with a gray background. 
The focus of this work is marked on the right: detecting mentioned claims in opponents speeches. Preceding existing components are shortly described in \sectionRef{sec:data}.
The entire pipeline starts from billions of sentences, and its final goal is producing few high quality rebuttals opposing the opponent speech.
\label{fig:claim-matching-architecture}}
\end{figure*}

\paragraph{Motions}

As in \citet{Mirkin-etal:idebate}, we manually curated a list of \numMotions controversial topics -– referred to as “motions”, as in formal parliamentary proposals. Each motion focuses on a single Wikipedia concept,
and is phrased similarly to parliamentary motions, e.g. \emph{We should introduce compulsory voting}. 

\paragraph{Speeches}
For each motion we recorded two argumentative speeches \emph{contesting} it, as described in \citet{Mirkin-etal:idebate}, producing a total of \numSpeeches speeches. 
Our choice of recording speeches contesting (rather than supporting) the motion  is arbitrary, and all methods described henceforth would work similarly on speeches recorded for the other side.
The dataset format follows the one described in \citet{Mirkin-etal:2018}.
Each speech is associated with a corresponding audio file, an automatic transcription of it\footnote{See details in \sectionRef{sec:eval}.}, and a manually-transcribed ``reference'' text. Speeches were recorded by \numDebaters expert debaters.
On average, a speech contains \averageNumberOfSentencePerSpeech sentences and  \averageNumberOfTokensPerSpeech tokens. 
The average ASR word error rate, computed by comparing to the manual transcripts, is \averageSpeechWER.

\paragraph{Mining Claims}  

\figRef{fig:claim-matching-architecture} illustrates the suggested mined--claims based rebuttal generation pipeline.
Following is a brief description of the existing components which perform claim mining. 
The rest of this work focuses on the subsequent component which identifies mentioned claims in speeches.

Processing starts from a large corpus of news articles containing billions of sentences. Given a controversial topic, several queries are applied, retrieving sentences which potentially contain claims that are relevant to the topic. Query results are then ranked by a neural-model trained to detect sentences containing claims (similarly to \newcite{Levy17Argmining,Levy18}\footnote{We note that, as opposed to cited work, the corpus used here is not Wikipedia.}). 
Top-ranked sentences are passed to a boundary detection component, responsible for finding the exact span of each claim within each sentence \cite{Levy14}.
Lastly, the stance of each claim towards the topic is detected using the method of \citet{Barhaim:2017}.
Used models are tuned towards precision, aimed at obtaining a set of coherent, grammatically--correct claims from the opponent side, which can then be directly quoted in a live debate.

Prior to claim matching, mined claims are filtered, aiming to focus on those with a higher chance of obtaining a successful match.
This included removing claims containing: (i) more than $10$ tokens, since longer claims are less concise and may contain more than a single idea; (ii) named entities (found with Stanford NER \cite{FinkelStanfordNER}), other than the topic itself, assuming 
they are too specific;
(iii) unresolved demonstratives, which may hint to an incoherent sentence or an error in boundary detection.

The released dataset includes all output from these components, as well as a 
complete labeling indicating which texts are erroneously predicted to be claims, and what is the correct stance of all valid claims. 
The percentage of mined texts which are both labeled as claims and have a correctly identified stance is \claimsPrecision.

\paragraph{Topic coverage}
Claim mining yielded, on average, \averageClaimsPerSpeech claims for each speech, suggesting match-candidates for \aceTopicCoveragePercentage of the motions in our data.
This shows the potentially high coverage of using mined claims.
In contrast, only \fractionOfMotionsInIdebate of these motions have candidate \idebate arguments present in the dataset of \citet{Mirkin-etal:idebate}.

\section{Annotation}\label{sec:labeling}

Next, we assessed whether mined claims are mentioned in recorded speeches through annotation.
In case mined claims do occur in many speeches, the collected labels would form a dataset which can be used to develop algorithms for identifying mined claims in speeches.

In our annotation scheme, each question included a speech followed by a list of mined claims, and 
we asked to mark those claims which were mentioned by the speaker.
Speeches were given in both text (manual transcription) and audio formats, to allow for listening, reading, or both.
The length of each claim list was limited to at most $20$ claims. Longer lists were split into multiple questions for the same speech.

Initially the task allowed for two labels: \emph{Mentioned} or \emph{Not mentioned}, 
yet error analysis showed major disagreements on 
claims alluded to, but not explicitly stated, in a speech.
Example~\ref{ex:pos-arg} illustrates this for the claim \emph{compulsory voting is undemocratic}.
Some 
annotators 
considered such cases as mentioned, while others disagreed.
Thus, we modified the task to include
three labels (\textit{Explicit}, \textit{Implicit}, \textit{Not mentioned}), and provided 
detailed examples in the guidelines.
Example~\ref{ex:pos-arg} further shows an explicit mention of the same claim\footnote{Full annotation guidelines, including more examples, are provided in the \supp.}.

\vspace{2mm}
\hrule
\begin{example}[Implicit / explicit mentions]\label{ex:pos-arg}
\emph{\textbf{\\Claim:} \textbf{\emph{Compulsory voting is undemocratic}}}
\vspace{1mm}
\emph{\textbf{\\Implicit} \emph{...people have a right to not vote ... that's the way that rights work ... if you think that there is literally any reason a person might not want to vote ... you should ensure that that person is not penalized for not voting...}}
\vspace{1mm}
\emph{\textbf{\\Explicit} \emph{...it might be preferable if everyone voted, but it is undemocratic to force everyone to vote.}}
\hrule
\end{example}

\paragraph{Quality control}

Annotation of each question is time-consuming, since it requires going over a whole speech, and a list of claims.
Combined with the amount of questions, 
we resorted to working with a crowd-sourcing platform\footnote{Figure-Eight: \url{www.figure-eight.com}}, to make annotation practical.
This required close monitoring and the removal of unreliable annotators.
For quality control, we placed ``test'' claims among real mined claims, either using claims from different motions, expecting a negative answer, or by using claims unanimously labeled as mentioned for the same speech in previous rounds, expecting a positive label (explicit or implicit).
We then defined thresholds on the accuracy of labeling of these test claims, and on the agreement of an annotator with its peers, disqualifying those who did not meet them.
In addition, good annotators were awarded bonus payments, in order to keep them engaged. 
Each question was answered by seven annotators. 

\paragraph{Annotation results} 
A claim is considered as \emph{mentioned} in a speech when a majority of annotators marked it as either an explicit or an implicit mention. 
A mentioned claim is an \emph{explicit} mention when its explicit answer count is strictly larger than its implicit answer count. Otherwise, it is an \emph{implicit} mention.

Overall, annotation of all $400$ speeches and their mined claims amounted to
\numLabeledSpeechLabelPairs speech--claim pairs. 
Of these, %$1,692$ 
$34.7\%$ 
were annotated as claims mentioned in the speech.
Only $5.6\%$ are explicit mentions, testifying to the difficulty of the matching task.

On average, there were $4.2$ mentioned claims in every speech.
$82.5\%$ of the labels 
were agreed on by at least $5$ out of the $7$ annotators. 
The percentage of claims mentioned at least once is $44.8\%$, and in 
%$349$ 
$87.3\%$ of speeches at least one claim is mentioned ($6.5\%$ of speeches had no mined claims).

\paragraph{Annotation Quality} 
To estimate inter-annotator agreement, we focus on annotators with a significant contribution, selecting 
those who have answered more than $20$ common questions with each of at least $5$ different peers.
A per-annotator agreement score is defined by averaging Cohen's Kappa \cite{cohenKappa} calculated with each peer.
The final agreement score is the average of all annotators agreement scores.

Considering two labels (mentioned or not), agreement was \cohenKappa. 
\citet{Mirkin-etal:idebate} reported a score of $0.5$ on a similar annotation scheme performed by expert annotators. 
The difference is potentially due to the use of crowd, and the larger group of annotators taking part.

Note the applicability of chance-adjusted agreement scores to the  crowd has been questioned, in particular for tasks within the argumentation domain \cite{passonneau2014benefits,habernal2016argument}.
Our test claims allow further validation of annotation quality, since their answers are known a-priory. The average annotator error rate on those test claims is low: $7.8\%$.

\section{Evaluation}\label{sec:eval}

Annotation confirmed our hypothesis that claims mined  from  a  corpus are indeed mentioned, or are at least alluded to,  in  spontaneous speeches on controversial topics.
On average, of the \averageClaimsPerSpeech claims mined for each speech, about a third were annotated as mentioned. 
We now present several baselines for identifying those mentioned claims, using the collected data. 

\paragraph{Speech pre--processing}
An input audio speech is automatically transcribed into text using IBM Watson ASR\footnote{\url{www.ibm.com/watson/services/speech-to-text}}.
The text is then segmented to sentences as in \citet{pahuja2017joint}.

Next, given a claim, semantically similar sentences are identified.
Each sentence is represented using a 200-dimensional vector constructed by: removing stopwords; representing remaining words using word2vec (\emph{w2v}) \cite{mikolov2013w2v} word embeddings learned over Wikipedia; computing a weighted average of those word embeddings using tf-idf weights (idf values are counted when considering each Wikipedia sentence as a document).
The claim is represented similarly, and its semantic similarity to a given sentence is computed using the cosine similarity between their vector representations. 
All sentences with low similarity to the claim are ignored (using a fixed threshold).

Remaining sentences are scored by 
the
harmonic mean (\hmName) of three additional semantic similarity measures, and the top-\K ranked sentence are selected (we experiment with  \K$\in\{1,3,5\}$). These features are: \\
\textit{\textbf{-- Concept Coverage}}: The fraction of Wikipedia concepts identified in the claim, found within the sentence. \\
\textit{\textbf{-- Parse Pairs}}: 
The parse trees of the claim and the sentence are obtained using Stanford parser \cite{socher2013parsing}. 
Then, pairwise edge similarity is defined to be the harmonic mean of the cosine similarities computed between the two parent word embeddings and the two child word embeddings.
Each edge in the claim parse tree is scored using its maximal similarity to an edge from the sentence parse tree.
Averaging these scores yields the final feature score. \\
\textit{\textbf{-- Explicit Semantic Analysis}}
\cite{Gabrilovich07esa}: Cosine similarity computed between vector representations of the claim and sentence over the Wikipedia concepts space. 

\paragraph{Methods}
Following sentence selection, three methods are considered for scoring a speech and a claim: \\
\listItem{HM}: Averaging the selected sentences \hmName scores. \\
\listItem{NN}: Using a Siamese Network \cite{Bromley:1993:SVU:2987189.2987282}, containing \K instances of the same sub-network: 
Each pair of a selected sentence and the claim is embedded with a BiLSTM, followed by an attention layer, a fully connected layer, and finally a \textit{softmax} layer which yields a score for the pair. The network outputs the maximum score of these \K sub-networks. \\
\listItem{LR}: calculating \numLogisticRegressionFeatures similarity measures 
between each selected sentence and the claim. For each measure, the average over the \K selected sentences is taken. These averages are used as features for training a logistic regression classifier.
Following is a brief description of the different groups of similarity measures we used.
\\
   \textbf{\textit{-- w2v-based similarities}} (5 features): 
    Computing pairwise word similarities using the cosine similarity of the corresponding word embeddings, and applying several aggregation options.
\\
      \textbf{\textit{-- Parse tree similarities}} (6 features): 
    Computing the parse tree of the claim and the sentence, and calculating similarities between different elements of those trees, similarly to  the \textit{Parse Pairs} feature described above.
\\
    \textbf{\textit{-- Part of speech (POS) similarities}} (5 features): 
    Identifying tokens with a specific POS tag in the texts, 
    and computing either the fraction of such tokens from one text which appear in the other, or otherwise aggregating w2v-based cosine similarities between these tokens in several ways.
    \\
    \textbf{\textit{-- Wikipedia concepts similarities}} (2 features): The fraction of Wikipedia concepts from the claim which are present in the sentence, and vice versa.
\\
    \textbf{\textit{-- Lexical similarities}} (5 features): 
    $n$-grams are extracted from the two texts in various settings (e.g. with or without lemmatization, or using different values of $n$). Then, each $n$-gram from the claim is scored by its maximal similarity to sentence $n$-grams (using a w2v-based similarity, with tf/idf weights). The feature values is the average of these scores.

\paragraph{Training and test sets}\label{sec:released}
The data was randomly split into a \devSetName and \testSetName sets, equal in size.
Each contains \splitSizeMotions motions and \splitSizeSpeeches speeches.
The number of labeled speech-claim pairs is \devNumSpeechClaimLabels in \devSetName and \testNumSpeechClaimLabels in \testSetName.

Model selection as well as hyper-parameters tuning, such as the selection of \K, are performed on \devSetName (using cross validation for \lrName and \nnName).
Different configuration are ranked according to their Area Under the ROC Curve (AUC) measure.  

\begin{figure}[t]
\centering
\includegraphics[width=75mm]{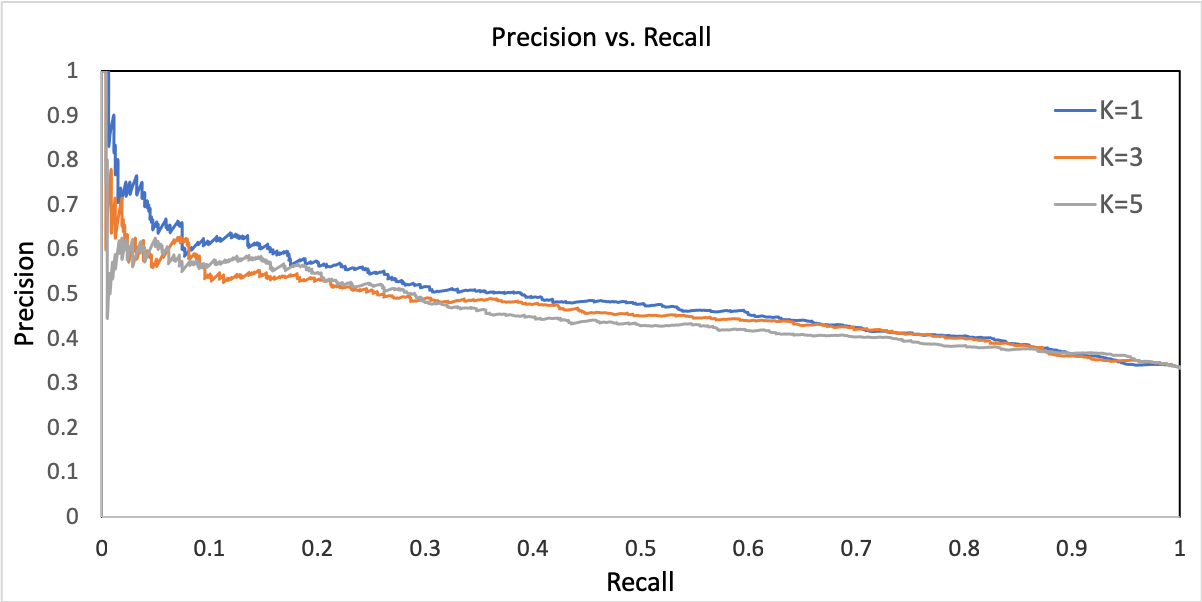}
\caption{Precision-Recall curves for the top-\numClaimMatchingMethodsSelectedForDisplay claim matching configurations (all using \hmName) on \testSetName.  \label{fig:precision-recall-test}}
\end{figure}

\paragraph{Results} The AUC score of both \lrName and \nnName on \devSetName, for various values of \K, was no higher than $0.57$.
In contrast, all \hmName configurations achieved AUC higher than $0.62$. 
We therefore focus on this  method, though it is interesting, in future work, to improve the supervised methods or understand why they work somewhat poorly. 
\figRef{fig:precision-recall-test} shows precision-recall curves for \hmName and the different values of \K on \testSetName.
The different plots are comparable, yet there is a slight advantage to \K$=1$ for applications valuing precision over recall.

\section{Conclusions and Future Work}
We addressed the task of identifying arguments claimed in spoken argumentative content.
Our suggested approach utilized claims mined from a large text corpora.
The collected labeled data show these claims do cover, in most cases, arguments made by expert debaters.
This confirms this is a valid approach for solving this task.

Interestingly, most claims are made implicitly, 
suggesting that assertion of claims often involves high lexical variability and expression of ideas across multiple (not always consecutive) sentences.
This poses a challenge for automatic claim matching methods, as made evident by the baselines discussed here.

Successfully identifying arguments made by opponents forms the basis for an effective rebuttal.
Our work leaves open the question of how to construct such rebuttals once a claim has been matched.
This would be an interesting research direction for future work.

\section{Acknowledgments}
We are thankful to the debaters and annotators who took part in the creation of this dataset. We thank George Taylor and the entire Figure-Eight team for their continuous support during the annotation process.

\bibliographystyle{styleAndBib/acl_natbib}
\bibliography{main}

\clearpage
\newpage
\appendix
\section{Annotation Guidelines}
\label{app:guidelines}

Following are the guidelines used in the annotation of mined claims to recorded speeches.
\vspace{3mm}

\noindent \textbf{Overview}

\noindent In the following task you are given a speech that contests a controversial topic. You are asked to listen to the speech and/or read the transcription, then decide whether a list of potentially related claims were mentioned by the speaker explicitly, implicitly, or not at all. 

\vspace{2mm}

\noindent \textbf{Steps}
\noindent \begin{itemize}

    \item \textbf{Listen} to the speech and/or read the transcription of the speech.
        Note: some speeches are transcribed automatically and may contain errors.
    \item \textbf{Review} the list of possibly relevant claims.
                Note: few of the claims might not be full sentences. Please do your best to ``complete'' them to claims in a common-sense manner. 
        If the claim doesn't make any sense, select ``Not mentioned''.
    \item \textbf{Decide} based on the speech only whether the speaker agrees with each claim, and choose the appropriate answer:
       \begin{itemize}
        \item Agree - Explicitly
        \item Agree - Implicitly
        \item Not Mentioned
    \end{itemize}
\end{itemize}

\noindent \textbf{Rules \& Tips}

\noindent You  should ask yourself whether the statement ``\emph{The speaker argued that \textless{}claim\textgreater{}}'' is valid or not. Note, this statement can be valid even if the speaker was stating the claim using a somewhat different phrasing in her/his  speech. 

\vspace{2mm}
\noindent \textbf{Examples}
  
\vspace{2mm}
\noindent  \textbf{\emph{Agree - Explicitly}}

\vspace{1mm}

\noindent The claim was mentioned by the speaker, but perhaps phrased differently.
    
           \begin{itemize}
        \item If the speaker said: \emph{organic food is simply healthier} then she explicitly agrees with the claim \textbf{organic food products are better in health}.
        \item If in a speech about the topic ``We should ban boxing'' the speaker said: \emph{we think regulation is simply better in this instance than a ban} then she explicitly agrees with the claim \textbf{We should not ban boxing altogether, just regulate it}.
       \end{itemize}

\noindent \textbf{\emph{Agree - Implicitly}}

\vspace{1mm}

\noindent The claim was not mentioned by the speaker but it is clearly implied from the speech, and we know for sure that the speaker agrees with the claim.
   
The claim will usually be implied in one of the following ways:
               \begin{itemize}

       \item  The claim is a generalization of a claim mentioned by the speaker. 
           
            If the speaker said: \emph{we allow people to make these decisions even if they might be physically bad for them} then she implicitly agrees with the claim \textbf{People should have the right to choose what to do with their bodies}.
        \item The claim summarizes an argument made by the speaker. 
            
            If the speaker said: \emph{It's essential that something is done to ensure that people don't have dental problems later in life. Water fluoridation is so cheap it's almost free. There are no proven side effects, the FDA and comparable groups in Europe have done lots and lots of tests and found that water fluoridation is actually a net health good, that there's no real risk to it} then she implicitly agrees with the claim \textbf{water fluoridation is safe and effective}.
       \item The claim can be deduced from an argument made by the speaker. 
            
            If the speaker said \emph{without the needle exchange program people are still going to do heroin or other kinds of drugs anyway with dirty or less safe needles. This does lead to things like HIV getting transmitted, it leads to other diseases as well, being more likely to get transmitted} then she implicitly agrees that \textbf{needle exchange programs could reduce the spread of disease}.
                       \end{itemize}

    The text itself must contain some indication of the implied claim. Don't choose this option if you need to make an extra logical step to conclude that the speaker agrees with the claim. For example, if the speaker said \emph{International aid has problems, but is still valuable}, then you should not conclude that she agrees with the claim \textbf{We should fix international aid, and not get rid of it} since she did not argue that the problems should be fixed.

\vspace{1mm}

\noindent \textbf{\emph{Not Mentioned}}

\vspace{1mm}

\noindent The claim is not part of the speech.
    
    For example, if the speaker said \emph{and, yes, feminism has its flaws in the status quo ... but it can be  reformed, and the tenets of equality that feminism stands for ... those  tenets certainly should not be abandoned, and feminism has done a  fantastic job, both historically and in the modern day, of championing  those tenets.} then it can not be inferred that she agrees with the claim \textbf{We should try to fix the issues with feminism because people support it}. Although she suggests to fix the issues with feminism, she does not claim that people support it.

\vspace{1mm}

\noindent \textbf{IMPORTANT NOTE}: Your answers will be reviewed after the job is complete. We trust you to perform the task thoroughly, while carefully following the guidelines. Once your answers are determined as acceptable per our review, you might receive a bonus. Note that the bonus is given to contributors who complete at least $5$ pages per job, and a higher bonus may be given to contributors who complete at least $50$ pages.

\end{document}